\documentclass[journal]{IEEEtran}

\usepackage{lipsum}
\usepackage{array}
\usepackage{times}
\usepackage{epsfig}
\usepackage{graphicx}
\usepackage{float}
\usepackage{wrapfig}
\usepackage{amsmath,amssymb,amsthm}
\usepackage{algorithm,algorithmicx,algpseudocode}
\usepackage{bm,xspace}
\usepackage{comment}
\usepackage{multirow}
\usepackage{balance}
\usepackage{url}
\usepackage{booktabs}
\usepackage{etoolbox,siunitx}
\usepackage{calc}
\usepackage{pifont,hologo}
\usepackage{color}
\usepackage{adjustbox}
\usepackage{enumitem}
\usepackage{bbding}  %
\usepackage[normalem]{ulem}  %
\usepackage{contour}
\usepackage[table]{xcolor}
\usepackage{soul}%
\usepackage{tocloft} %
\usepackage{etoc} %
\PassOptionsToPackage{dvipsnames}{xcolor}
\usepackage[breaklinks,colorlinks,linkcolor=link,citecolor=cite]{hyperref} % pagebackref
\usepackage[caption=false,font=normalsize,labelfont=sf,textfont=sf]{subfig}
\usepackage{mathtools}
\usepackage{orcidlink}

\usepackage{fancyhdr}

\definecolor{blue}{HTML}{004bb3}
\definecolor{red}{HTML}{cc1100}
\definecolor{orange}{HTML}{cc7700}
\definecolor{gray}{HTML}{efefef}
\definecolor{darkgreen}{HTML}{228B22}
\definecolor{darkgray}{HTML}{757575}

\definecolor{cite}{HTML}{3270b5}
\definecolor{link}{HTML}{b53532}
\definecolor{link}{HTML}{cc1100}
\definecolor{scratch}{HTML}{001219}
\definecolor{pretrain}{HTML}{0A9396}
\definecolor{sm}{HTML}{001219}
\definecolor{smpro}{HTML}{36B0A4}
\definecolor{smap50}{HTML}{1f77b4}

\definecolor{fe}{HTML}{ECEDFE}
\definecolor{qi}{HTML}{EAF2FA}
\definecolor{qr}{HTML}{A9EAF9}

\definecolor{m_pink}{RGB}{253, 244, 244}
\definecolor{m_blue}{RGB}{234, 242, 250}
\definecolor{m_gray}{RGB}{242, 242, 244}

\definecolor{backbone}{RGB}{220, 234, 247}
\definecolor{saqs}{RGB}{234, 242, 250}
\definecolor{psqd}{RGB}{202, 238, 251}

\newcommand{\sm}{\textcolor{sm}{$\mathbf{\circ}$\,}}
\newcommand{\smpro}{\textcolor{smpro}{$\bullet$\,}}

\contourlength{0.8pt}

\newcommand{\figref}[1]{Fig.~\ref{#1}}
\newcommand{\tabref}[1]{Tab.~\ref{#1}}
\newcommand{\secref}[1]{Sec.~\ref{#1}}
\renewcommand{\eqref}[1]{Eq.~\ref{#1}}

\newcolumntype{x}[1]{>{\centering\arraybackslash}p{#1}}
\newcolumntype{y}[1]{>{\raggedright\arraybackslash}p{#1}}
\newcolumntype{z}[1]{>{\raggedleft\arraybackslash}p{#1}}

\DeclareMathSymbol{@}{\mathord}{letters}{"3B}

\newcommand\mypara[1]{\vspace{0mm}\noindent\textbf{#1}}

\newcommand{\YesV}{\ding{51}}%

\makeatletter
\DeclareRobustCommand\onedot{\futurelet\@let@token\@onedot}
\def\@onedot{\ifx\@let@token.\else.\null\fi\xspace}
 
\def\ie{\emph{i.e}\onedot} 
 
 \def\vs{\emph{vs}\onedot}

\newcommand*{\Rom}[1]{\expandafter\@slowromancap\romannumeral #1@}
\newcommand*{\rom}[1]{\expandafter\romannumeral #1}

\def\1{\bm{1}}

\def\rvp{{\mathbf{p}}}

\def\rvw{{\mathbf{w}}}

\def\rmA{{\mathbf{A}}}
\def\rmB{{\mathbf{B}}}
\def\rmC{{\mathbf{C}}}
\def\rmD{{\mathbf{D}}}

\def\rmF{{\mathbf{F}}}

\def\rmI{{\mathbf{I}}}

\def\rmK{{\mathbf{K}}}

\def\rmP{{\mathbf{P}}}
\def\rmQ{{\mathbf{Q}}}

\def\rmS{{\mathbf{S}}}

\def\rmV{{\mathbf{V}}}
\def\rmW{{\mathbf{W}}}
\def\rmX{{\mathbf{X}}}

\def\vc{{\bm{c}}}

\def\vm{{\bm{m}}}

\def\vs{{\bm{s}}}

\DeclareMathAlphabet{\mathsfit}{\encodingdefault}{\sfdefault}{m}{sl}
\SetMathAlphabet{\mathsfit}{bold}{\encodingdefault}{\sfdefault}{bx}{n}

\def\gC{{\mathcal{C}}}
\def\gD{{\mathcal{D}}}

\def\gJ{{\mathcal{J}}}

\def\gL{{\mathcal{L}}}
\def\gM{{\mathcal{M}}}

\newcommand{\R}{\mathbb{R}}

\let\originalleft\left
\let\originalright\right
\renewcommand{\left}{\mathopen{}\mathclose\bgroup\originalleft}
\renewcommand{\right}{\aftergroup\egroup\originalright}

\newcommand{\ours}{LaSSM\xspace}
\newcommand{\scannet}{ScanNet V2\xspace}
\newcommand{\scannetp}{ScanNet200\xspace}
\newcommand{\sdis}{S3DIS\xspace}
\newcommand{\scannetpp}{ScanNet++\xspace}
\newcommand{\spformer}{SPFormer~\cite{sun2023spformer}\xspace}

\begin{document}

\title{\ours: Efficient Semantic-Spatial Query Decoding via Local Aggregation and State Space Models for 3D Instance Segmentation}

\author{Lei Yao,
        Yi Wang,
        Yawen Cui,
        Moyun Liu,
        and~Lap-Pui Chau% <-this % stops a space
        
\thanks{\emph{Corresponding author: Lap-Pui Chau}}

\thanks{Lei Yao, Yi Wang, Yawen Cui, and Lap-Pui Chau are with the Department of Electrical and Electronic Engineering, The Hong Kong Polytechnic University, Hong Kong.}% <-this % stops a space

\thanks{Moyun Liu is with the School of Mechanical Science and Engineering, Huazhong University of Science and Technology, Wuhan 430074, China.}
}

\maketitle

\begin{abstract}
    Query-based 3D scene instance segmentation from point clouds has attained notable performance. However, existing methods suffer from the query initialization dilemma due to the sparse nature of point clouds and rely on computationally intensive attention mechanisms in query decoders. We accordingly introduce \textit{\ours}, prioritizing simplicity and efficiency while maintaining competitive performance. Specifically, we propose a hierarchical semantic-spatial query initializer to derive the query set from superpoints by considering both semantic cues and spatial distribution, achieving comprehensive scene coverage and accelerated convergence. We further present a coordinate-guided state space model (SSM) decoder that progressively refines queries. The novel decoder features a local aggregation scheme that restricts the model to focus on geometrically coherent regions and a spatial dual-path SSM block to capture underlying dependencies within the query set by integrating associated coordinates information. Our design enables efficient instance prediction, avoiding the incorporation of noisy information and reducing redundant computation. \ours ranks \textit{first place} on the latest \scannetpp V2 leaderboard, outperforming the previous best method by 2.5\% mAP with only 1/3 FLOPs, demonstrating its superiority in challenging large-scale scene instance segmentation. \ours also achieves competitive performance on \scannet, \scannetp, \sdis and \scannetpp V1 benchmarks with less computational cost. Extensive ablation studies and qualitative results validate the effectiveness of our design. The code and weights are available at \href{https://github.com/RayYoh/LaSSM}{https://github.com/RayYoh/LaSSM}.
\end{abstract}


\section{Introduction}
\label{sec:intro}
\IEEEPARstart{P}{erceiving} and comprehending 3D environments constitutes a fundamental requirement for embodied agents, including intelligent robots~\cite{zheng2024survey} and AR/VR systems~\cite{wirth2019pointatme}, since these application scenarios require precise understanding of spatial configurations and object-level semantics~\cite{yao2025gaussian}. Within this framework, point cloud 3D instance segmentation emerges as a critical task, entailing the simultaneous prediction of instance masks and associated semantic labels in given scenes.

\begin{figure}[t]
    \centering
    \includegraphics[width=0.48\textwidth]{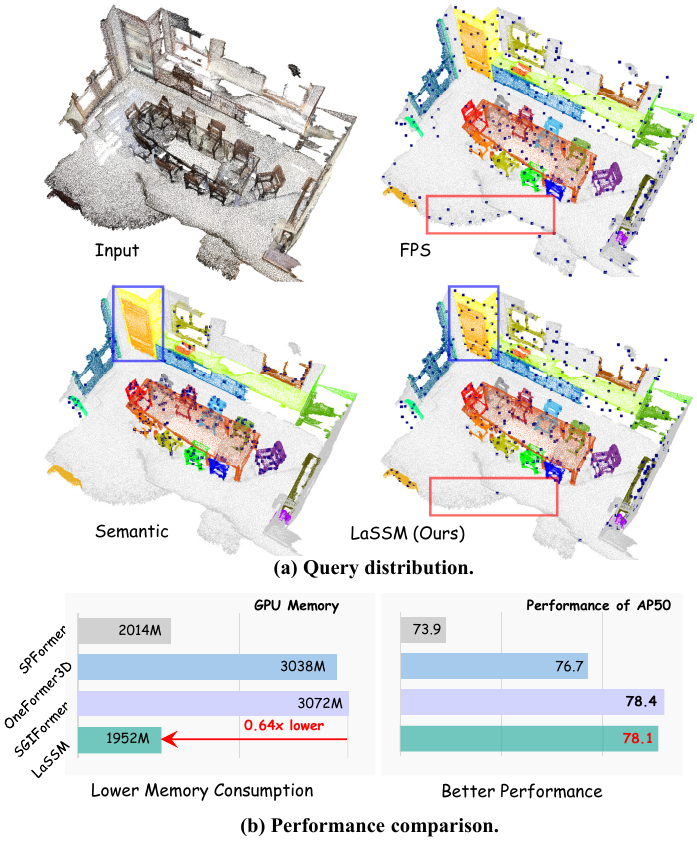}
    \caption{\textbf{Query distribution and performance comparison.} (a) We compare query distributions of farthest point sampling (FPS)~\cite{schult2023mask3d}, semantic confidence-based selection (Semantic)~\cite{he2023fastinst}, and our method on different scenes. (b) Compared to SPFormer~\cite{sun2023spformer}, OneFormer3D~\cite{kolodiazhnyi2023of3d} and SGIFormer~\cite{yao2024SGIFormer}, \ours achieves the balance between performance and GPU efficiency.}
    \label{fig:teaser}
\end{figure}

Recent advances in query-based 3D instance segmentation~\cite{schult2023mask3d, sun2023spformer, lu2023query, lai2023mask, kolodiazhnyi2023of3d, yao2024SGIFormer, lu2025beyond, lu2025relation3d} have demonstrated superior performance by leveraging the end-to-end set prediction paradigm, which utilizes a transformer decoder casting a set of queries to instances. However, two fundamental challenges remain inadequately addressed: (1) \textbf{\textit{optimal query initialization}} given sparse point distributions and multi-scale object variations, and (2) \textbf{\textit{effective yet efficient query refinement}} avoiding computational redundancy. Current initialization strategies exhibit distinct limitations: learnable queries~\cite{sun2023spformer} usually lack adaptability to varying scene complexities, while geometry-based sampling methods~\cite{schult2023mask3d} frequently select non-discriminative positions. For example, Mask3D's farthest point sampling (FPS) implementation allocates many queries to non-instance regions (\figref{fig:teaser}a, \textcolor{red}{red box}). Semantic confidence selection~\cite{he2023fastinst} introduces spatial bias, failing to recognize some instances(\textcolor{blue}{blue box} in~\figref{fig:teaser}a).

Equally critical is the computational efficiency during query refinement. The standard practice of applying cross-attention~\cite{cheng2022mask2former} across all decoder layers introduces quadratic complexity with query quantity, resulting in redundant computations in typical indoor scenes (\figref{fig:teaser}b). This scalability bottleneck becomes particularly prohibitive when processing large-scale environments. In addition, the lack of positional information in the query refinement stage can lead to suboptimal instance localization, as observed in~\cite{yao2024SGIFormer}. This limitation hinders the overall performance, especially in complex scenes with intricate object arrangements.

In this work, we investigate a novel query-based 3D instance segmentation framework, \textbf{\ours}, to alleviate the aforementioned challenges. Our architecture consists of two key components: an optimized query initializer for high-quality query construction and an efficient query decoder for accurate instance prediction. Specifically, the proposed \textit{hierarchical semantic-spatial query initializer} adaptively determines query contents and associated coordinates from superpoints~\cite{felzenszwalb2004efficient}, considering both semantic cues and spatial distribution. This mechanism ensures thorough coverage of scenes (\figref{fig:teaser}a), while significantly accelerating the convergence process (\figref{fig:query_convergence}). Moreover, the \textit{coordinate-guided state space model-based query decoder} refines instance queries with linear complexity. Each of decoder layers comprises a local aggregation scheme and a spatial dual-path state space model (SSM) block. Given the inherent sparsity of point clouds where instances typically occupy a small portion of the scene~\cite{wang2023long}, we first introduce the local aggregation to accumulate geometric proximity features into query contents, enhancing their representation. This technique enforces queries to focus on coherent regions, thereby avoiding redundant computation and noisy information. As discussed in~\cite{yao2024SGIFormer}, prior query-based methods generally ignore positional details~\cite{lin2023dbganet}, leading to suboptimal query localization. Although SGIFormer~\cite{yao2024SGIFormer} introduces position encoding to mitigate this issue, latent embeddings will lose critical geometric relationships. In contrast, our spatial dual-path SSM block directly sorts the query set into regular sequences according to Hilbert space curve~\cite{hilbert1935stetige} based on their coordinates. However, this position-aware sequencing will be disrupted by transformer decoding since its invariance to input order~\cite{vaswani2017attention}. Thus, we propose to utilize the capability of SSM~\cite{gu2023mamba} to capture potential dependencies and facilitate communication among queries, bypassing the computational overhead of attention mechanisms~\cite{vaswani2017attention}. Concurrently, a dedicated center regression module updates query coordinates to align with content refinements. 

Extensive evaluations demonstrate the superiority of \ours across various benchmarks. \ours ranks \textbf{\textit{first place}} on challenging \scannetpp V2 leaderboard at submission time, surpassing the previous state-of-the-art method by 2.5$\%$ mAP and 2.3$\%$ AP$_{50}$ while requiring only 1/3 of FLOPs. It also achieves competitive results on \scannet~\cite{dai2017scannet}, \scannetp~\cite{hou2021csc}, and \scannetpp V1~\cite{yeshwanth2023scannet++} with substantially reduced computational cost. This achievement underscores the effectiveness of our method in balancing performance and efficiency. Ablation studies systematically validate the impact of each component. In summary, our contributions are as follows:
\begin{itemize}
    \item We introduce \ours, a novel framework for query-based 3D instance segmentation.
    \item We propose a hierarchical semantic-spatial query initializer that adaptively constructs the high-quality query set from superpoints, ensuring comprehensive scene coverage and accelerated convergence.
    \item We develop a coordinate-guided SSM query decoder, including a local aggregation scheme and a spatial dual-path SSM block, achieving efficient query refinement and positional information incorporation.
    \item \ours ranks \textbf{\textit{first place}} on the \scannetpp V2 leaderboard and demonstrates competitive performance on \scannet, \scannetp, \sdis and \scannetpp V1 benchmarks with less computational cost, establishing its practical value for large-scale 3D scene understanding.
\end{itemize}

\section{Related Work}
\label{sec:related}

\subsection{Query-based 3D Instance Segmentation}
Current 3D instance segmentation methods are broadly classified into three paradigms: proposal-based~\cite{yang2019learning, kolodiazhnyi2023td3d}, cluster-based~\cite{jiang2020pointgroup, chen2021hais, vu2022softgroup, liang2021sstnet}, and query-based approaches~\cite{schult2023mask3d, sun2023spformer, lu2023query, lai2023mask, al20233d, kolodiazhnyi2023of3d, yao2024SGIFormer, lu2025beyond, lu2025relation3d}. Proposal-based frameworks adopt a top-down strategy by first generating coarse object proposals and refining them through subsequent optimization stages. Cluster-based methods operate bottom-up, learning point-level embeddings and aggregating them into instances based on semantic and geometric cues. However, both paradigms suffer from performance bottlenecks due to error propagation in intermediate processing stages. The emergence of query-based approaches, inspired by DETR's set prediction paradigm~\cite{carion2020detr,su2025annexe}, has shifted the paradigm toward direct instance prediction. Recent works, such as Mask3D~\cite{schult2023mask3d} and \spformer, directly predicted instances via transformer decoders to progressively refine queries. In particular, SPFormer relied entirely on learnable queries, whereas Mask3D employed nonparametric FPS~\cite{qi2017pointnet++} and used zero-initialized features as query contents. QueryFormer~\cite{lu2023query} further enhanced the performance by introducing a query aggregation module to improve spatial coverage at the cost of increased computational complexity. OneFormer3D~\cite{kolodiazhnyi2023of3d} proposed a stochastic query selection strategy during training, but reverting to exhaustive superpoint utilization during inference. BFL~\cite{lu2025beyond} proposed an agent-interpolation query initialization integrating FPS with learnable queries and a query fusion transformer decoder, while Relation3D~\cite{lu2025relation3d} introduced a relation-aware self-attention mechanism to enhance query interactions. However, both methods still face challenges in terms of efficiency.

Differently, our approach introduces a hierarchical semantic-spatial query initializer, which synergistically leverages semantic confidence while ensuring spatial coverage.

\begin{figure*}[!htbp]
    \centering
    \includegraphics[width=0.98\textwidth]{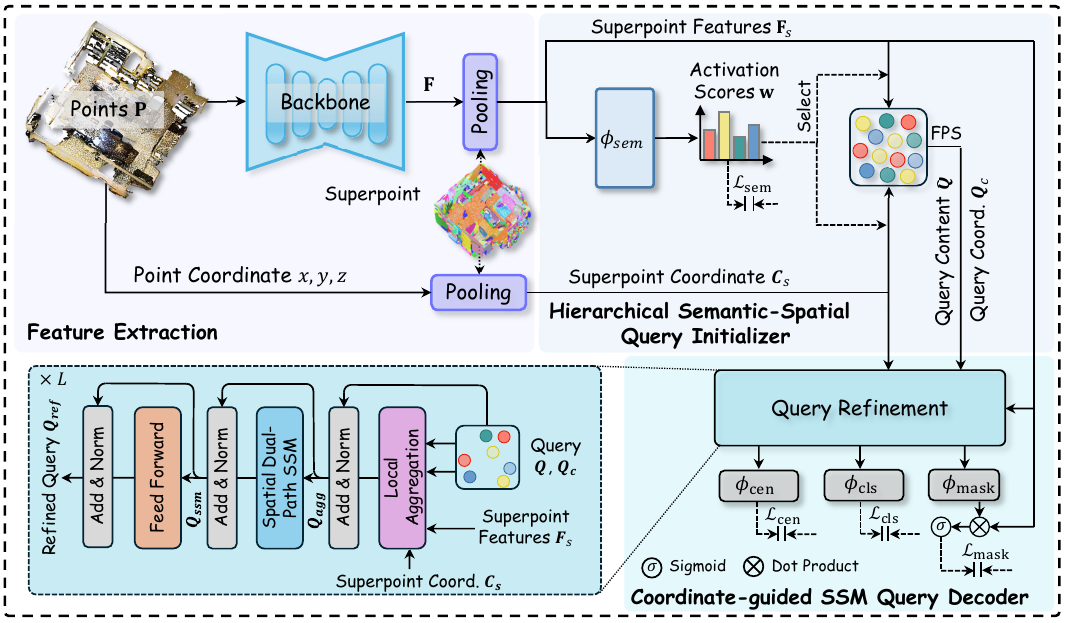}
    \caption{\textbf{Architecture of \ours.} The input point cloud is processed by the feature extractor to obtain superpoint features $\rmF_s$ and coordinates $\rmC_s$. Then the hierarchical semantic-spatial initializer is employed to initialize query contents $\rmQ$ and coordinates $\rmQ_c$ (\secref{subsec:saqs}). The resulting query set is further refined by the coordinate-guided SSM query decoder, which iteratively updates query contents and coordinates to predict instances (\secref{subsec:pgqd}). FPS denotes farthest point sampling.}
    \label{fig:architecture}
\end{figure*}

\subsection{Efficient Query Decoders}
The quadratic complexity inherent in attention mechanisms~\cite{vaswani2017attention} has driven the development of efficient alternatives, such as linear attention~\cite {vaswani2017attention}, to reduce computational overhead. AdaMixer~\cite{gao2022adamixer} replaced transformer layers with adaptive MLP-Mixers~\cite{tolstikhin2021mlp}, explicitly modeling channel-wise semantics and spatial relationships through lightweight operations. PEM~\cite{cavagnero2024pem} introduced an efficient cross-attention that selects the most relevant features as prototypes for each query.  BoxeR~\cite{nguyen2022boxer} restricted attention to bounding box regions predicted from queries, while SEED~\cite{liu2024seed} further built upon this concept by deformable grid attention that predicts query-specific spatial offsets. To address depth stacking inefficiencies, SGIFormer~\cite{yao2024SGIFormer} developed an interleaved decoder architecture that alternates between global and local feature interactions, reducing dependency on deep layer cascading. 

In contrast, \ours proposes a coordinate-guided query decoder consisting of a local aggregation scheme and a spatial dual-path SSM block.

\subsection{State Space Models in Computer Vision}
SSMs, originally developed for modeling dynamical systems, have recently emerged as competitive alternatives to transformers in sequential data processing. Mamba~\cite{gu2023mamba} first introduced a selection mechanism to ignore irrelevant information while maintaining hardware-aware computational efficiency. Vision adaptations such as ViM~\cite{zhu2024vision, cao2025m} extended this capability to images by bidirectional scanning strategies, demonstrating superior performance in dense prediction tasks. Inspired by its success, a series of methods~\cite{zhang2024point, liang2024pointmamba, wang2024pointramba, han2024mamba3d, zhang2024voxel} try to adapt SSMs to unstructured point clouds. Typically, these methods divide the whole pipeline into two phases: serializing points into structured tokens and utilizing SSMs to capture inherent information. For example, Point Mamba~\cite{liu2024seed} organized points via octree-based ordering combined with ViM-style bidirectional processing. PCM~\cite{zhang2024point} and PointMamba~\cite{liang2024pointmamba} explored integrating Z-order~\cite{morton1966computer} and Hilbert-order~\cite{hilbert1935stetige} serialization mechanisms. Voxel Mamba~\cite{zhang2024voxel} proposed a dual-scale SSM block to improve spatial proximity. 

While existing SSM-based methods primarily focus on object-level classification or point-wise semantic segmentation, our work pioneers the integration of SSMs into query decoding for 3D scene instance segmentation.

\section{Methodology}
\label{sec:method}
This section begins with a brief introduction of SSMs and introduces our architectural framework as depicted in~\figref{fig:architecture}. We then delve into the details of our hierarchical semantic-spatial query initializer and the tailored coordinate-guided SSM-based query decoder.

\subsection{Preliminaries}
\label{subsec:preliminaries}
SSMs are initially designed to model the dynamics of continuous systems in control theory by hidden state evolution. Given a sequence of input states $x(t) \in \R^{L}$ and hidden states $h(t) \in \R^{N}$, SSMs map to outputs $y(t) \in \R^{L}$ via:
\begin{equation}
    h^{\prime}(t) = \rmA h(t) + \rmB x(t), \quad
    y(t) = \rmC h(t) + \rmD x(t),
    \label{eq:ssm}
\end{equation}
where $h^{\prime}$ is the derivative of $h$, $\rmA \in \R^{N \times N}$ denotes the state transition matrix, $\rmB \in \R^{N \times L}$ represents the input matrix describing the effect of input on hidden states, while $\rmC \in \R^{L \times N}$ and $\rmD \in \R^{L \times L}$ are projection matrix and residual connection, respectively. Most implementations omit the $\rmD$ term, \ie, $\rmD x(t) = 0$. To integrate SSMs into deep learning architectures, zero-order hold (ZOH) discretization is applied to convert the continuous system into a discrete one using a time scale parameter $\Delta$. This allows to rewrite \eqref{eq:ssm} as: 
\begin{equation}
        h_{k} = \overline{\rmA} h_{k-1} + \overline{\rmB} x_{k}, \quad
        y_{k} = \rmC h_{k},
    \label{eq:ssm_discrete}
\end{equation}
where $\overline{\rmA} = e^{\Delta \rmA}, \overline{\rmB} = (\Delta \rmA)^{-1}(e^{\Delta \rmA} - \rmI) \cdot \Delta \rmB$. Here, $x_{k}, h_{k}$, and $y_{k}$ are discrete inputs, hidden states, and outputs at time step $k$, and $\rmI$ indicates the identity matrix. This formulation enables efficient convolutional computation~\cite{gu2023mamba} by:
\begin{equation}
        \overline{\rmK} = (\rmC \overline{\rmB}, \rmC \overline{\rmA} \overline{\rmB}, \ldots, \rmC \overline{\rmA}^{L-1} \overline{\rmB}), \quad
        y = x * \overline{\rmK},
\end{equation}
where $L$ is the sequence length and $\overline{\rmK} \in \R^L$ represents the convolution kernel. Furthermore, \cite{dao2024transformers} introduces the concept of state space duality, replacing matrix $\rmA$ with a scalar to enhance flexibility. Our model is built upon this variant.

\subsection{Overall Architecture}
\label{subsec:overall}
As illustrated in~\figref{fig:architecture}, our method comprises three primary components: a sparse 3D U-Net backbone~\cite{graham2017submanifold}, a hierarchical semantic-spatial query initializer, and a coordinate-guided SSM query decoder. By taking the voxelized scene point cloud $\rmP \in \R^{n \times 6}$, which includes $x,y,z$ coordinates and $r,g,b$ colors, as input, the sparse symmetrical backbone is responsible for extracting voxel-wise features $\rmF = \texttt{Backbone}(\rmP) \in \R^{n \times d_o}$, where $n$ denotes voxel count and $d_o$ represents channel dimension. We then apply a superpoint pooling operation~\cite{felzenszwalb2004efficient} after the feature extractor to cluster voxels into superpoints:
\begin{equation}
    \rmF_s = \texttt{SpPool}(\rmF), \quad
    \rmC_s = \texttt{SpPool}\left([x,y,z]\right),
    \label{eq:avgpool}
\end{equation}
yielding superpoint features $\rmF_s \in \R^{s \times d_o}$ and coordinates $\rmC_s \in \R^{s \times 3}$, where $s$ is superpoint number. Subsequently, the initializer is designed to efficiently initialize a set of non-learnable instance queries $\rmQ \in \R^{q \times d}$, where $q$ and $d$ are query numbers and feature dimensions. The decoder is further utilized to cast queries into potential instances by progressively refining them.

\subsection{Hierarchical Semantic-spatial Query Initializer}
\label{subsec:saqs}
The quality of query initialization critically influences the efficacy of 3D instance segmentation methods. Mask3D~\cite{schult2023mask3d} has demonstrated the advantages of non-learnable queries, which zero-initialize query contents for sampled points by FPS. OneFormer3D~\cite{kolodiazhnyi2023of3d} proposed a query selection strategy to pick a subset of queries from superpoints randomly, but discarded this mechanism during inference to prioritize segmentation performance. In this work, we introduce a hierarchical semantic-spatial query initializer to initialize instance queries and retain the selection mechanism during both training and inference stages. Thus, the model prioritizes superpoints with high semantic confidence and spatial distribution, mitigating the query initialization issue discussed in~\secref{sec:intro}.

Specifically, a lightweight MLP classifier $\phi_\text{sem}$ is leveraged on superpoint features $\rmF_s$ to yield category probabilities, formulated as $\rmS = \phi_\text{sem}(\rmF_s) \in \R^{s \times (c+1)}$, where $c$ denotes instance categories and the extra dimension corresponds to background. We further apply a \textit{softmax} normalization on the output to get semantic activation scores $\rvw_i \in \R^{c+1}$ per superpoint $i$. To prioritize instance-specific regions, we extract the maximum non-background confidence by:
\begin{equation}
    w_i^* = \underset{j \in \gJ}{\max}\{w_{i,j} | w_{i,j} \in \rvw_{i}\},
    \label{eq:scores}
\end{equation}
where $\gJ = \{1, 2, \ldots, c\}$. We then rank superpoints based on predicted activation scores and select the top-$m$ superpoints for further processing. Instead of setting a static threshold, we dynamically adjust the number of candidates according to the input scene. We exploit a ratio $r$ to control the selected number. This adaptive thresholding allows the model to adjust candidate selection according to scene complexity, and this step can be formulated as:
\begin{equation}
    m = \lfloor r \cdot s \rfloor, \quad
    \mathbf{ids}_s = \texttt{TopK}(\{{w_i^*}\}_{i=0}^s, m),
\end{equation}
where $\mathbf{ids}_s$ is the selected superpoint indices. Although our concise strategy can filter out irrelevant regions, it may result in duplicate queries. Therefore, we further apply FPS on the selected superpoint features to sample $q$ queries, ensuring spatial coverage. The final query contents are obtained by projecting selected features into high-dimensional embeddings:
\begin{equation}
    \mathbf{ids}_f = \texttt{FPS}(\rmF_s[\mathbf{ids}_s], q), \quad
    \rmQ = \phi_\text{proj}(\rmF_s[\mathbf{ids}_f]),
\end{equation}
where $\phi_\text{proj}$ is implemented by simple MLP and $\rmQ \in \R^{q \times d}$. Corresponding query coordinates $\rmQ_{c} = \rmC_s[\mathbf{ids}_f] \in \R^{q \times 3}$ are simultaneously generated for subsequent decoding stages.

\begin{algorithm}[!htbp]
    \caption{Local Aggregation}
    \begin{algorithmic}[1]
        \Require query contents and coordinates $\rmQ$, $\rmQ_{c}$, number $k$, \Statex \quad \quad {~~~}superpoint features and coordinates $\rmF_{s}$, $\rmC_{s}$, \Statex \quad \quad {~~} learnable weight matrix $\rmK$
        \Ensure aggregated queries $\rmQ_\text{agg}$
        \State \textcolor{darkgray}{\text{/* get representative features by $k$-NN */}}
        \State $\rmV \in \R^{q \times k \times d} \gets \texttt{KNN}(\rmF_{s}, \rmC_{s}, \rmQ_{c}, k)$
        \State \textcolor{darkgray}{\text{/* expand query contents to align with $\rmV$ */}}
        \State $\rmQ' \in \R^{q \times k \times d} \gets \texttt{Expand}(\rmQ, k)$
        \State \textcolor{darkgray}{\text{/* enhance interaction */}}
        \State $\rmX_i \in \R^{d} \gets \sum_{j=1}^k \rmK_{i,j} \cdot (\rmQ'_{i,j} \odot \rmV_{i,j})$
        \State $\rmX \in \R^{q \times d} \gets \{\rmX_i\}_{i=1}^q$
        \State \textcolor{darkgray}{\text{/* project to original space and residual connection */}}
        \State $\rmQ_\text{agg} \in \R^{q \times d} \gets \rmQ + \rmW_o \cdot \rmX$
        \State \textbf{Return}: $\rmQ_\text{agg}$
    \end{algorithmic}
    \label{alg:aggregation}
\end{algorithm}

\subsection{Coordinate-guided SSM Query Decoder}
\label{subsec:pgqd}
The detailed structure of our decoder is illustrated in~\figref{fig:architecture}. By treating queries $\rmQ$ as a token sequence, the query refinement processes them by $L$ stacked layers. It consists of a \textit{local aggregation} scheme to attend features from geometrically relevant regions, and a \textit{spatial dual-path SSM} block to facilitate intra-query communication by incorporating positional information. The prediction module is responsible for predicting instance masks and corresponding categories.

\mypara{Local Aggregation.}
Unlike dense image regions, 3D instances typically occupy sparse, localized portions of scenes. Although masked cross-attention partially constrains queries to focus on candidate regions, %%its quadratic computational complexity persists, while 
imperfect mask prediction may introduce irrelevant noise. Building upon recent efficient decoder designs~\cite{nguyen2022boxer, cavagnero2024pem, gao2022adamixer}, we propose a geometry-constrained local aggregation scheme that restricts query interactions to spatially coherent superpoints, as detailed in Algorithm~\ref{alg:aggregation}. For each query coordinate $\rmQ_{c}$, we perform $k$-nearest neighbor search over superpoint coordinates $\rmC_s$ to identify spatially proximate candidates and consider corresponding embeddings $\rmV \in \R^{q \times k \times d}$ as representative features. To model positional relevance, each query is associated with a learnable importance vector $K_i \in \R^{k}$ by a normalized \textit{softmax} function. These weights collectively form a matrix $\rmK = \{K_i\}_{i=1}^q$, which modulates channel-wise interactions between query contents $\rmQ' \in \R^{q \times k \times d}$ (expanded from $\rmQ$) and $\rmV$ through an element-wise product, resulting in $\rmX_i \in \R^{d}$ for each query $i$:
\begin{equation}
    \rmX_i = \sum_{j=1}^k \rmK_{i,j} \cdot (\rmQ'_{i,j} \odot \rmV_{i,j}).
\end{equation}
This operation efficiently enhances the interaction in both spatial and channel dimensions. Finally, we project intermediate results back to the original query contents space using a linear transformation $\rmW_o \in \R^{d \times d}$ with a residual connection to obtain the aggregated query contents $\rmQ_\text{agg} \in \R^{q \times d}$. In this way, we make the model concentrate on coherent superpoint regions constrained by the spatial proximity, reducing the input dimension from $s \times d$ to $k \times d$ ($k \ll s$).

\begin{figure}[t]
    \centering
    \includegraphics[width=0.49\textwidth]{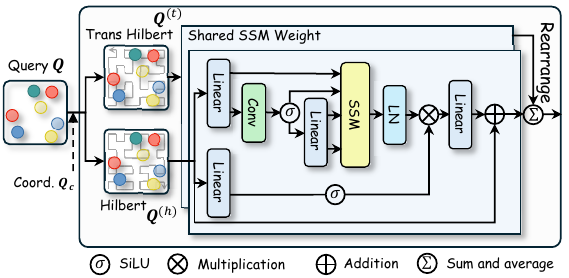}
    \caption{\textbf{Detailed architecture of spatial dual-path SSM block.}}
    \label{fig:ssm}
\end{figure}

\mypara{Spatial Dual-path SSM Block.}
The local aggregation is followed by a spatial dual-path SSM block to incorporate positional information and enhance intra-query communication, as shown in \figref{fig:ssm}. Previous transformer-based decoders~\cite{sun2023spformer, kolodiazhnyi2023of3d} ignore significant positional details and leverage permutation invariance of self-attention to model query interactions, which leads to suboptimal instance localization as discussed in~\cite{yao2024SGIFormer}. In contrast, our method emphasizes the importance of positional relationships by serializing unordered queries into a structured sequence $\rmQ^{(h)}$, maintaining local geometric continuity. Then, we consider SSMs to capture underlying patterns, leveraging the inherent causal modeling nature. However, since the unidirectional modeling ability of SSMs, we construct a complementary sequence $\rmQ^{(t)}$ using the transposed Hilbert curve proposed in~\cite{wu2024point}. Both sequences are fed into a shared SSM block, and use the mean to update query contents:
\begin{equation}
    \rmQ_\text{ssm} = \rmQ_\text{agg} + \frac{1}{|\gD|} \sum_{\rmQ^{(i)}_{\text{agg}} \in \gD} \texttt{RA}(\texttt{SSM}(\rmQ^{(i)}_{\text{agg}})),
\end{equation}
where $\gD = \{\rmQ^{(h)}_{\text{agg}}, \rmQ^{(t)}_{\text{agg}}\}$ denotes dual-path inputs, and $\texttt{RA}(\cdot)$ rearranges outputs back to original order using pre-recorded permutation indices. Final refinement is achieved through residual-connected feed-forward networks:
\begin{equation}
    \rmQ_\text{ref} = \rmQ_\text{ssm} + \texttt{FFN}(\rmQ_\text{ssm}).
\end{equation}
This dual-path design preserves the spatial coherence and mitigates directional bias.

\mypara{Instance Prediction Head.} Given refined queries $\rmQ_\text{ref}$ from each decoder layer, we predict instance masks $\gM$ and categories $\rvp$ using an instance prediction module:
\begin{equation}
    \gM = \left\{\sigma \left( \rmF_s \cdot \phi_\text{mask}(\rmQ_\text{ref})^\top \right)_{i,j} > \tau \right\},
    \label{eq:mask}
\end{equation}
\begin{equation}
    \rvp = \phi_\text{cls}(\rmQ_\text{ref}),
\end{equation}
where $\sigma(\cdot)$ denotes the \textit{sigmoid} function, $\tau$ is a threshold, and 
$\phi_\text{mask}, \phi_\text{cls}$ denote mask and classification heads implemented by separate MLP. To maintain positional accuracy for iterative refinement, we simultaneously predict instance center offsets by $\rmQ_{c}' = \rmQ_{c} + \phi_\text{cen}(\rmQ_\text{ref})$, where $\phi_\text{cen}$ is an MLP-based adjustment head. This technique enhances spatial precision in subsequent local aggregation and SSM operations.

\subsection{Training and Inference}
\label{subsec:train}
With available semantic annotations, we derive superpoint-wise category labels $\vs^* \in \R^{s}$ by majority voting. The semantic activation module is supervised via cross-entropy loss:
\begin{equation}
    \gL_\text{sem} = \frac{1}{n}\sum_{i=1}^n \texttt{CE}(\vs_i, \vs^*_i),
\end{equation}
where $\vs_i$ represents predicted semantic scores. Following standard practice~\cite{schult2023mask3d,sun2023spformer,yao2024SGIFormer}, we use bipartite graph matching~\cite{karp1990optimal} to assign associated instance pairs at each decoder layer. The matching cost between the $i$-th predicted instance and the $j$-th ground truth combines multiple objectives:
\begin{equation}
    \begin{split}
    \gC_{i, j} = &\lambda_\text{mask} \left(\texttt{BCE}(\gM_i, \gM^{*}_{j}) + \texttt{Dice}(\gM_i, \gM^{*}_{j})\right) \\
        & + \lambda_\text{cls} \texttt{CE} (\rvp_i, \rvp^{*}_{j}) +\lambda_\text{cen} \lVert \rmQ_{c,i}, \rmQ^{*}_{c,j}\rVert_1,
    \end{split}
\end{equation}
where $\gM^{*}_{j}$, $\rvp^{*}_{j}$, and $\rmQ^{*}_{c,j}$ represent ground truth masks, categories, and instance centers, respectively. The Hungarian algorithm~\cite{kuhn1955hungarian} determines optimal assignments by minimizing matching costs. The overall loss function is formulated as:
\begin{equation}
    \begin{split}
        \gL = & \sum_{l=1}^{L} \left[ \lambda_\text{cls} \gL^l_\text{cls} +\lambda_\text{cen} \gL^l_\text{cen} + \lambda_\text{mask} (\gL^l_\text{bce} + \gL^l_\text{dice}) \right] \\
        & + \lambda_\text{sem}\gL_\text{sem},
    \end{split}
    \label{eq:loss}
\end{equation}
with balancing coefficients $\lambda_\text{cls} = 0.5,$ $\lambda_\text{mask} = 1.0$, $\lambda_\text{cen} = 1.0$ and $\lambda_\text{sem} = 0.2$

During inference, the model processes raw point clouds to directly output $I$ instances with predicted masks $\{\vm_i\}_{i=1}^I$ and category scores $\{\vc_i\}_{i=1}^I$. For each mask, we calculate a confidence score $\gamma_i = \vc_i \cdot z_i$, where $z_i$ is the average superpoint probability within the mask region exceeding the threshold. Final point-wise predictions are obtained by mapping superpoint masks to original points.
\section{Experiments}

\begin{table}[t]
    \centering
    \caption{\textbf{Performance comparison on \scannetpp V2 benchmark.} Results are reported on 07/05/2025\protect\footnotemark. The best results are shown in \textbf{bold}, and the second best are \underline{underlined}. \sm denotes the model adopts our query refinement decoder, while~\smpro indicates the hybrid model.}
    \resizebox{0.49\textwidth}{!}{
    \begin{tabular}{y{20mm}x{5mm}x{5mm}x{5mm}x{5mm}x{5mm}x{5mm}}
        \toprule
        \multirow{2}{*}{\textbf{\scannetpp V2}} & \multicolumn{3}{c}{\textbf{\textit{Val}}} & \multicolumn{3}{c}{\textbf{\textit{Test}}} \\
        \cmidrule(lr){2-4} \cmidrule(lr){2-4} \cmidrule(lr){5-7}
        {} & mAP &AP$_{50}$ & AP$_{25}$ & mAP &AP$_{50}$ & AP$_{25}$ \\
        \midrule
        PointGroup~\cite{jiang2020pointgroup} & - & 14.7 & - & 9.1 & 15.2 & 21.0\\
        SoftGroup~\cite{vu2022softgroup} & - & - & - & 14.2 & 23.4 & 29.3 \\
        PTv3-PG~\cite{wu2024point} & - & - & - & 25.9 & 38.7 & 44.6 \\
        OneFormer3D~\cite{kolodiazhnyi2023of3d} & - & 41.1 & - & 28.2 & 43.3 & 51.7 \\
        SPFormer~\cite{sun2023spformer} & - & \underline{42.1} & - & 29.0 & 43.5 & 51.5 \\
        SGIFormer~\cite{yao2024SGIFormer} & - & 41.1 & - & \underline{29.9} & \underline{45.7} & \underline{54.4} \\
        \midrule
        \rowcolor{gray} \sm \ours(Ours) & \underline{27.6} & 41.9 & \underline{51.2} & - & - & - \\
        \rowcolor{gray} \smpro \ours(Ours) & \textbf{29.1} & \textbf{43.5} & \textbf{51.6} & \textbf{32.4} & \textbf{48.0} & \textbf{54.8} \\
        \bottomrule
    \end{tabular}
}
    \label{tab:scannetppv2}
\end{table}
\footnotetext{The results are obtained from \href{https://github.com/scannetpp/scannetpp}{official benchmark} and \href{https://kaldir.vc.in.tum.de/scannetpp/benchmark/insseg}{evaluation server}.}

\mypara{Datasets and Evaluation Metrics.} We evaluate \ours on several commonly used indoor datasets: \scannetpp V2\footnote{See ScanNet++ V2 \href{https://kaldir.vc.in.tum.de/scannetpp/}{webpage.}}, V1~\cite{yeshwanth2023scannet++}, \scannet~\cite{dai2017scannet}, \scannetp~\cite{hou2021csc} and \sdis~\cite{armeni2016s3dis}. \textit{\scannetpp V1} is a large-scale and challenging indoor dataset with sub-millimeter resolution and complex scenes, making it more realistic for real-world applications. The dataset is divided into training, validation, and testing sets of 230, 50, and 50 scans, respectively, containing 84 categories for 3D instance segmentation. \textit{\scannetpp V2} is a recently released benchmark that includes notable improvements to V1, including 856, 50, and 50 scans for training, validation, and testing. \textit{\scannet} includes 1201 and 300 annotated indoor scenes for training and validation, respectively, covering 18 instance categories. As an extension, \textit{\scannetp} provides more fine-grained semantic labels, with 198 object classes for instance segmentation. \textit{\sdis}~\cite{armeni2016s3dis} dataset consists of 6 different areas with 271 scanned rooms, which are annotated by 13 instance categories. Following previous work~\cite{sun2023spformer,yao2024SGIFormer}, our evaluation metrics include standard average precision mAP, AP$_{50}$, AP$_{25}$.

\mypara{Implementation Details.} Our model is implemented using Pointcept~\cite{pointcept2023} based on PyTorch and is trained for 512 epochs employing an AdamW optimizer with a weight decay of 0.05. The initial learning rate is set to $3e^{-4}$, with a polynomial learning rate scheduler leveraging a base value of 0.9. The models are trained on 4 NVIDIA RTX 4090 GPUs with a batch size of 12. In terms of data augmentation, we apply a series of transformation methods following SGIFormer~\cite{yao2024SGIFormer} to the original point cloud input. For model settings, we set the number of stacked layers in the decoder to 6 and the query numbers to 400 for all datasets during the training and inference stages. We leverage Mamba-2~\cite{dao2024transformers} as the implementation for our spatial dual-path SSM block, and set the number of hidden sizes to 256 and state size to 64.

\begin{table}[t]
    \centering
    \caption{\textbf{Performance comparison on \scannetpp V1 benchmark.} Results on the testing set are accessed on 05/11/2024.}
    \resizebox{0.49\textwidth}{!}{
    \begin{tabular}{y{19mm}x{6mm}x{4mm}x{4mm}x{4mm}x{4mm}x{4mm}x{4mm}}
        \toprule
        \multirow{2}{*}{\textbf{\scannetpp V1}} & \multirow{2}{*}{\textbf{Year}} & \multicolumn{3}{c}{\textbf{\textit{Val}}} & \multicolumn{3}{c}{\textbf{\textit{Test}}} \\
        \cmidrule(lr){3-5} \cmidrule(lr){6-8}
        {} & {} & mAP &AP$_{50}$ & AP$_{25}$ & mAP &AP$_{50}$ & AP$_{25}$ \\
        \midrule
        PointGroup \cite{jiang2020pointgroup} & 2020 & - & 14.8 & - & 8.9 & 14.6 & 21.0\\
        HAIS \cite{chen2021hais} & 2021 & - & 16.7 & - & 12.1 & 19.9 & 29.5 \\
        SoftGroup \cite{vu2022softgroup} & 2022 & - & 23.7 & - & 16.7 & 29.7 & 38.9 \\
        MAFT~\cite{lai2023mask} & 2023 & \underline{23.1} & 32.6 & 39.7 & 20.9 & 31.9 & 41.3 \\
        BFL~\cite{lu2025beyond} & 2025 & \textbf{25.3} & \underline{35.2} & 42.6 & 22.2 & 32.8 & 42.5 \\
        Relation3D~\cite{lu2025relation3d} & 2025 & - & - & - & \underline{24.2} & \underline{35.5} & \underline{44.0} \\
        \midrule
        \rowcolor{gray} \sm \ours(Ours) & - & 21.0 & 34.6 & \underline{43.8} & - & - & - \\
        \rowcolor{gray} \smpro \ours(Ours) & - & 21.3 & \textbf{35.8} & \textbf{44.8} & \textbf{25.2} & \textbf{37.7} & \textbf{47.1} \\
        \bottomrule
    \end{tabular}
}
    \label{tab:scannetpp}
\end{table}

\subsection{Comparison with State-of-the-art Methods}

In our results, the marker~\sm denotes the model adopts our proposed decoder, while~\smpro indicates a hybrid decoder which integrates masked cross-attention in the first 3 layers and our local aggregation scheme in the remaining layers. {We consider this hybrid decoder to balance the performance and efficiency.

\begin{table}[t]
    \centering
    \caption{\textbf{Performance comparison on \scannet benchmark.} $\dagger$ We reproduce OneFormer3D results, decreasing top instances to 200 during inference for fair comparison.}
    \resizebox{0.49\textwidth}{!}{
    \begin{tabular}{y{22mm}x{15mm}x{6mm}x{6mm}x{6mm}x{6mm}}
        \toprule
        \multirow{2}{*}{Methods} & \multirow{2}{*}{Venue} & \multicolumn{3}{c}{Val} & Test \\
        \cmidrule(lr){3-5} \cmidrule(lr){6-6}
        {} & {} & mAP &AP$_{50}$ & AP$_{25}$ & mAP \\
        \midrule
        \multicolumn{6}{c}{\textbf{\textit{Proposal-based}}} \\
        \midrule
        3D-SIS~\cite{hou20193dsis} & CVPR 19 & - & 18.7 & 35.7 & 16.1 \\
        GSPN~\cite{yi2019gspn} & CVPR 19 & 19.3 & 37.8  & 53.4 & - \\
        TD3D~\cite{kolodiazhnyi2023td3d} & WACV 24 & 47.3 & 71.2 & 81.9 & 47.3 \\
        \midrule
        \multicolumn{6}{c}{\textbf{\textit{Cluster-based}}} \\
        \midrule
        JSNet++ \cite{zhao2022jsnet++} & TCSVT 22 & - & 39.2 & 56.8 & - \\
        PointGroup~\cite{jiang2020pointgroup} & CVPR 20 & 34.8 & 56.9 & 71.3 & 40.7 \\
        SSTNet~\cite{liang2021sstnet} & ICCV 21 & 49.4 & 64.3 & 74.0 & 50.6 \\
        HAIS~\cite{chen2021hais} & ICCV 21 & 43.5 & 64.4 & 74.6 & 45.7 \\
        SoftGroup~\cite{vu2022softgroup} & CVPR 22 & 45.8 & 67.6 & 78.9 & 50.4 \\
        PBNet~\cite{zhao2023pbnet} & ICCV 23 & 54.3 & 70.5 & 78.9 & 57.3 \\
        ISBNet~\cite{ngo2023isbnet} & CVPR 23 & 54.5 & 73.1 & 82.5 & 55.9 \\
        \midrule
        \multicolumn{6}{c}{\textbf{\textit{Query-based}}} \\
        \midrule
        Mask3D \cite{schult2023mask3d} & ICRA 23 & 55.2 & 73.7 & 82.9 & 56.6 \\
        \spformer & AAAI 23 & 56.3 & 73.9 & 82.9 & 54.9 \\
        3IS-ESSS \cite{al20233d} & ICCV 23 & 56.1 & 75.0 & 83.7 & - \\
        OneFormer3D$^\dagger$ \cite{kolodiazhnyi2023of3d} & CVPR 24 & 57.5 & 76.3 & 84.0 & 56.6 \\
        SGIFormer~\cite{yao2024SGIFormer} & TCSVT 25 & \textbf{58.9} & \textbf{78.4} & \textbf{86.2} & \textbf{58.6} \\
        \midrule
        \rowcolor{gray} \sm \ours (Ours) & - & 57.6 & 76.8 & 85.2 & - \\
        \rowcolor{gray} \smpro \ours (Ours) & - & \underline{58.4} & \underline{78.1} & \underline{86.1} & \underline{57.9} \\
        \bottomrule
    \end{tabular}
}
    \label{tab:scannet_val}
\end{table}

\mypara{\scannetpp V1 and V2.} As depicted in~\tabref{tab:scannetppv2} and~\tabref{tab:scannetpp}, we benchmark \ours against existing methods on \scannetpp V2 and V1 to verify the feasibility of processing large-scale indoor scenes. In~\tabref{tab:scannetppv2}, \smpro \ours establishes a new state-of-the-art performance on both validation and test splits, outperforming previous best results by 1.4\% and 2.3\% in terms of AP$_{50}$, respectively. Notably, \sm \ours even gets slightly better accuracy than SGIFormer~\cite{yao2024SGIFormer}. In~\tabref{tab:scannetpp}, it is obvious that two versions of our model consistently exceed previous methods, and \smpro \ours achieves better AP$_{50}$ and AP$_{25}$ with 35.8\% and 44.8\% on the validation set. Although our mAP is lower than MAFT~\cite{lai2023mask} and BFL~\cite{lu2025beyond}, results on the hidden test set are significantly better, and surpass Relation3D~\cite{lu2025relation3d} by a large margin. In~\tabref{tab:efficiency} \textit{right}, even though \smpro \ours has a slightly larger model size and slower inference speed, we demonstrate that it attains the best efficiency with only 1/3 of average FLOPs compared to SGIFormer~\cite{kolodiazhnyi2023of3d}, which reveals its superiority when handling high-resolution point clouds with complex layouts and diverse object categories.

\mypara{\scannet.} In~\tabref{tab:scannet_val}, we compare \ours with existing counterparts on \scannet benchmark. On the validation split, \sm \ours outperforms most of prior methods, including SPFormer~\cite{sun2023spformer} by 1.3\% mAP. \smpro \ours further improves mAP to 58.4\% and 57.9\% on validation and test sets, surpassing OneFormer3D~\cite{kolodiazhnyi2023of3d}. Due to the sparse instance distribution of \scannet, our local aggregation scheme may introduce some background noise since the utilization of KNN, leading to a slight performance gap compared to SGIFormer~\cite{yao2024SGIFormer}. However, our method achieves a significant reduction in average FLOPs (4.456G \textit{v.s.} 3.711G) for each scene, as shown in~\tabref{tab:efficiency}, which indicates that \ours can achieve competitive results with lower computational cost. Furthermore, we observe that query-based methods generally perform better than proposal-based and cluster-based methods, which highlights its advancement in handling error propagation issue.

\begin{table}[t]
    \centering
    \caption{\textbf{Performance comparison on \scannetp validation set.}}
    
\resizebox{0.49\textwidth}{!}{
    \begin{tabular}{y{25mm}x{15mm}x{8mm}x{8mm}x{8mm}}
        \toprule
        Methods & Venue & mAP &AP$_{50}$ & AP$_{25}$ \\
        \midrule
        SPFormer \cite{sun2023spformer} & AAAI 23 & 25.2 & 33.8 & 39.6 \\
        TD3D \cite{kolodiazhnyi2023td3d} & WACV 24 & 23.1 & 34.8 & 40.4 \\
        MSTA3D \cite{tran2024msta3d} & MM 24 & 26.2 & 35.2 & 40.1 \\
        Mask3D \cite{schult2023mask3d} & ICRA 23 & 27.4 & 37.0 & 42.3 \\
        QueryFormer \cite{lu2023query} & ICCV 23 & 28.1 & 37.1 & 43.4 \\
        MAFT \cite{lai2023mask} & ICCV 23 & \underline{29.2} & 38.2 & 43.3 \\
        SGIFormer~\cite{yao2024SGIFormer} & TCSVT 25 & \underline{29.2} & \textbf{39.4} & \underline{44.2} \\
        \midrule
        \rowcolor{gray} \sm \ours (Ours) & - & 28.8 & 38.2 & 43.7 \\
        \rowcolor{gray} \smpro \ours (Ours) & - & \textbf{29.3} & \underline{39.2} & \textbf{44.5} \\
        \bottomrule
    \end{tabular}
}
    \label{tab:scannetp}
\end{table}

\begin{table}[t]
    \centering
    \caption{\textbf{Performance comparison on \sdis Area5.} $^*$ means the results of pretrained on \scannet.}
    
\resizebox{0.45\textwidth}{!}{
    \begin{tabular}{y{25mm}x{15mm}x{8mm}x{8mm}}
        \toprule
        Methods &Venue & mAP & AP$_{50}$ \\
        \midrule
        PointGroup \cite{jiang2020pointgroup} & CVPR 20 & - & 57.8 \\
        TD3D \cite{kolodiazhnyi2023td3d} & WACV 24 & 48.6 & 65.1 \\
        SoftGroup \cite{vu2022softgroup} & CVPR 22 & 51.6 & 66.1 \\
        PBNet \cite{zhao2023pbnet} & ICCV 23 & 53.5 & 66.4 \\
        SPFormer \cite{sun2023spformer} & AAAI 23 & - & 66.8 \\
        OneFormer3D$^*$ \cite{kolodiazhnyi2023of3d} & CVPR 24 & - & 68.5 \\
        MAFT \cite{lai2023mask} & ICCV 23 & - & 69.1 \\
        SGIFormer~\cite{yao2024SGIFormer} & TCSVT 25 & \underline{56.2} & \underline{69.2} \\
        \midrule
        \rowcolor{gray} \smpro \ours (Ours) & - & \textbf{56.5} & \textbf{69.4} \\
        \bottomrule
    \end{tabular}
}
    \label{tab:s3dis}
\end{table}

\mypara{\scannetp.} We evaluate our method on \scannetp for more fine-grained 3D instance segmentation in~\tabref{tab:scannetp}. The table demonstrates that our approach can exceed most of the existing methods. Specifically, \sm \ours achieves similar accuracy to MAFT~\cite{lai2023mask} with 38.2\% AP$_{50}$ and 43.7\% AP$_{25}$, while \smpro \ours obtains competitive results with SGIFormer~\cite{yao2024SGIFormer} with 39.2\% AP$_{50}$ and 44.5\% AP$_{25}$. These findings further underscore the effectiveness of our method in handling more fine-grained instance categories and long-tail distributions. 

\mypara{\sdis.} As shown in~\tabref{tab:s3dis}, \smpro \ours also obtains competitive results on \sdis Area5, achieving 56.5\% mAP and 69.4\% AP$_{50}$, which are still better SGIFormer~\cite{yao2024SGIFormer}. This further demonstrates the superiority of our method in transferring to different indoor datasets.

\begin{table}[t]
    \centering
    \caption{\textbf{Efficiency comparison.} We compare AP$_{50}$ and average FLOPs of \ours with OneFormer3D~\cite{kolodiazhnyi2023of3d} and SGIFormer~\cite{yao2024SGIFormer}.}
    
\resizebox{0.49\textwidth}{!}{
    \begin{tabular}{y{19mm}x{3mm}x{8mm}x{3mm}x{7mm}x{7mm}x{9mm}}
        \toprule
        \multirow{2}{*}{Methods} & \multicolumn{2}{c}{\textbf{ScanNet V2}} & \multicolumn{4}{c}{\textbf{\scannetpp V2}} \\
        \cmidrule(lr){2-3} \cmidrule(lr){4-7} 
        {} & AP$_{50}$ & FLOPs$\downarrow$ & AP$_{50}$ & FLOPs$\downarrow$ & Params$\downarrow$ & Time$\downarrow$ \\
        \midrule
        OneFormer3D~\cite{kolodiazhnyi2023of3d} & 76.3 & 4.082G & 41.1 & OOM & - & - \\
        SGIFormer~\cite{yao2024SGIFormer} & \textbf{78.4} & 4.456G & 41.1 & 13.513G & \textbf{15.96M} & \textbf{355.86ms} \\
        \midrule
        \rowcolor{gray} \smpro \ours (Ours) & 78.1 & \textbf{3.711G} & \textbf{43.5} & \textbf{4.777G} & 18.38M & 379.94ms \\
        \bottomrule
    \end{tabular}
}
    \label{tab:efficiency}
\end{table}

\begin{figure}[t]
    \centering
    \includegraphics[width=0.47\textwidth]{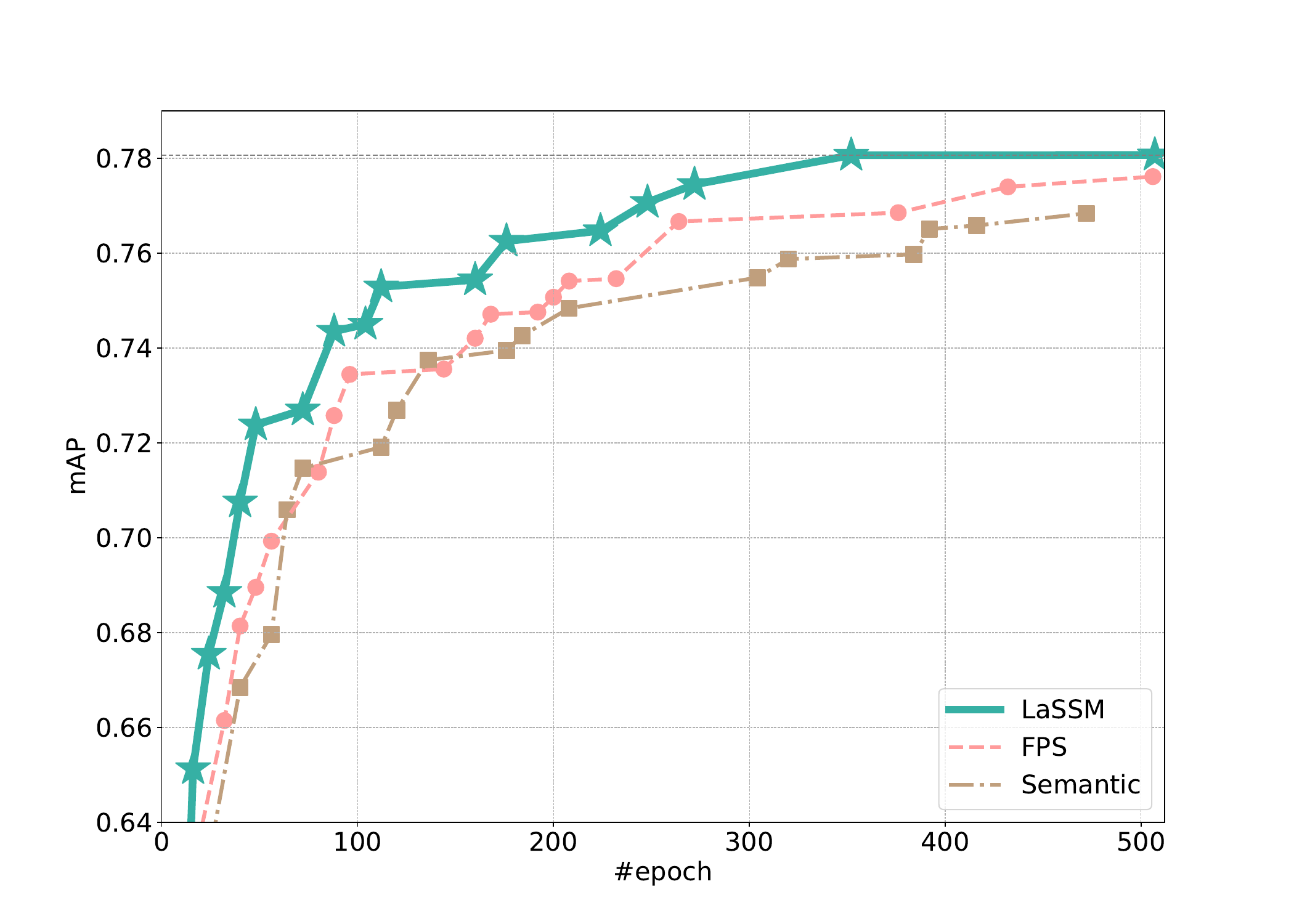}
    \caption{\textbf{Convergence speed.} \ours enables faster convergence and better performance than farthest point sampling (FPS)~\cite{schult2023mask3d} and semantic-guided~\cite{he2023fastinst} query selection methods.}
    \label{fig:query_convergence}
\end{figure}

\begin{table}[t]
    \centering
    \caption{\textbf{Query initialization methods and efficient query refinement decoders.}}

\resizebox{0.49\textwidth}{!}{
    \begin{tabular}{x{14mm}x{18mm}x{4mm}x{4mm}x{4mm}x{10mm}}
        \toprule
        Item & Type & mAP & AP$_{50}$ & AP$_{25}$ & Memory$\downarrow$ \\
        \midrule
        \multirow{2}{*}{\shortstack{\textit{Query}\\\textit{Initialization}}} & FPS~\cite{schult2023mask3d} & 57.8 & 77.6 & 84.8 & 1978M \\
        {} & Semantic~\cite{he2023fastinst} & 57.7 & 76.8 & 84.9 & 2004M \\
        \midrule
        \multirow{4}{*}{\textit{\shortstack{\textit{Query}\\\textit{Decoding}}}} & AdaMixer~\cite{gao2022adamixer} & 57.6 & 77.4 & 85.6 & 2020M \\
        {} & KVT~\cite{wang2022kvt} & 56.9 & 77.8 & 85.5 & 1988M \\
        {} & PEM~\cite{cavagnero2024pem} & 58.1 & 77.1 & 84.9 & 2198M \\
        {} & Lin. Attn.~\cite{katharopoulos2020transformers} & 58.0 & 77.2 & 85.1 & 2172M \\
        \midrule
        \rowcolor{gray} \multicolumn{2}{c}{\smpro Ours} & \textbf{58.4} & \textbf{78.1} & \textbf{86.1} & \textbf{1952M} \\
        \bottomrule
    \end{tabular}
}
    \label{tab:query_decoder}
\end{table}

\begin{table}[t]
    \centering
    \caption{\textbf{Effectiveness of core components in our decoder.}}
    \resizebox{0.49\textwidth}{!}{
    \begin{tabular}{x{7mm}|x{5mm}x{5mm}|x{5mm}x{7mm}|x{5mm}x{5mm}x{5mm}x{11mm}}
        \toprule
        \multirow{2}{*}{\#} & \multirow{2}{*}{\shortstack{Cross\\Attn.}} & \multirow{2}{*}{\shortstack{Local\\Agg.}} & \multirow{2}{*}{\shortstack{Self\\Attn.}} & \multirow{2}{*}{\shortstack{Spatial\\SSM}} & \multirow{2}{*}{mAP} & \multirow{2}{*}{AP$_{50}$} & \multirow{2}{*}{AP$_{25}$} & \multirow{2}{*}{Memory $\downarrow$} \\
        {} & {} & {} & {} & {} & {} & {} \\
        \midrule
        1 & \YesV & {} & \YesV & {} & 57.1 & 77.3 & 85.0 & 2020M \\
        2 & \YesV & {} & {} & \YesV & 58.3 & 77.7 & 85.2 & 2008M \\
        \midrule
        \rowcolor{gray} \smpro Ours & {} & \YesV & {} & \YesV & 58.4 & 78.1 & 86.1 & \textbf{1952M} \\
        \bottomrule
    \end{tabular}
}
    \label{tab:components}
\end{table}

\begin{figure*}[!htbp]
    \centering
    \includegraphics[width=0.96\textwidth]{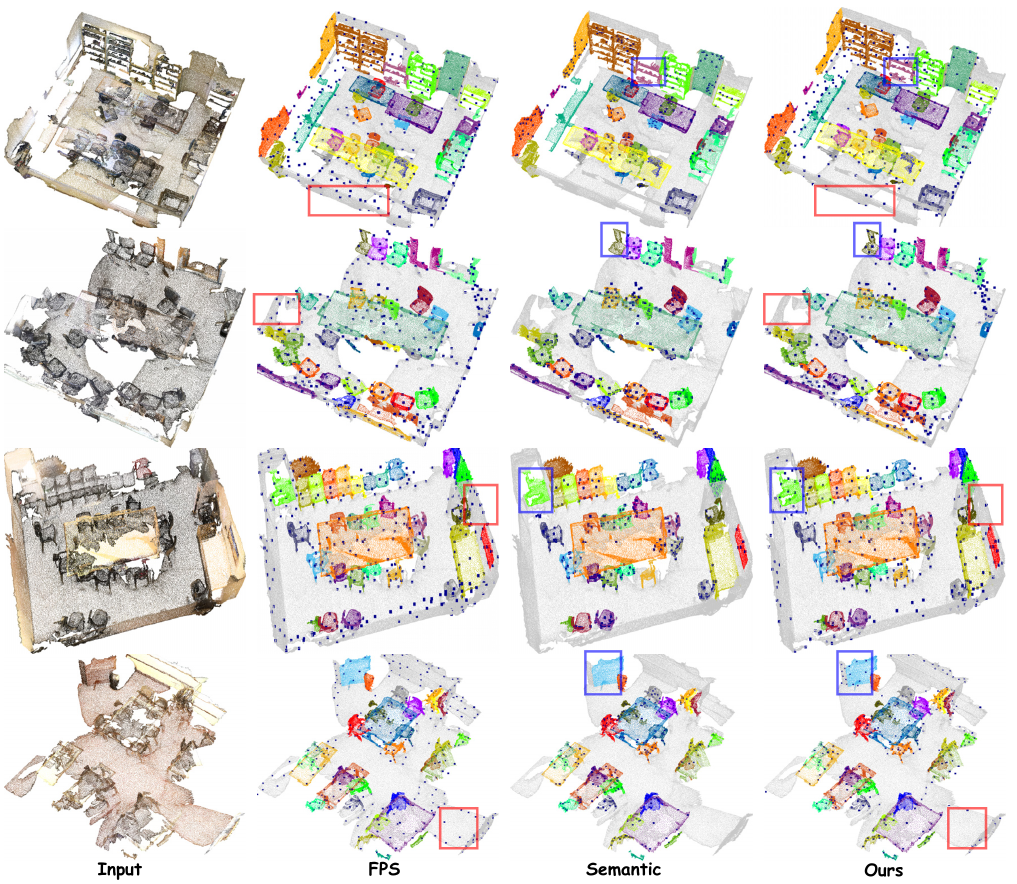}
    \caption{\textbf{Visualization of query distributions.} We emphasize the comparison of FPS and Semantic in the \textcolor{red}{red} and \textcolor{blue}{blue} boxes, respectively.}
    \label{fig:viz_query}
\end{figure*}

\begin{figure*}[!htbp]
    \centering
    \includegraphics[width=0.98\textwidth]{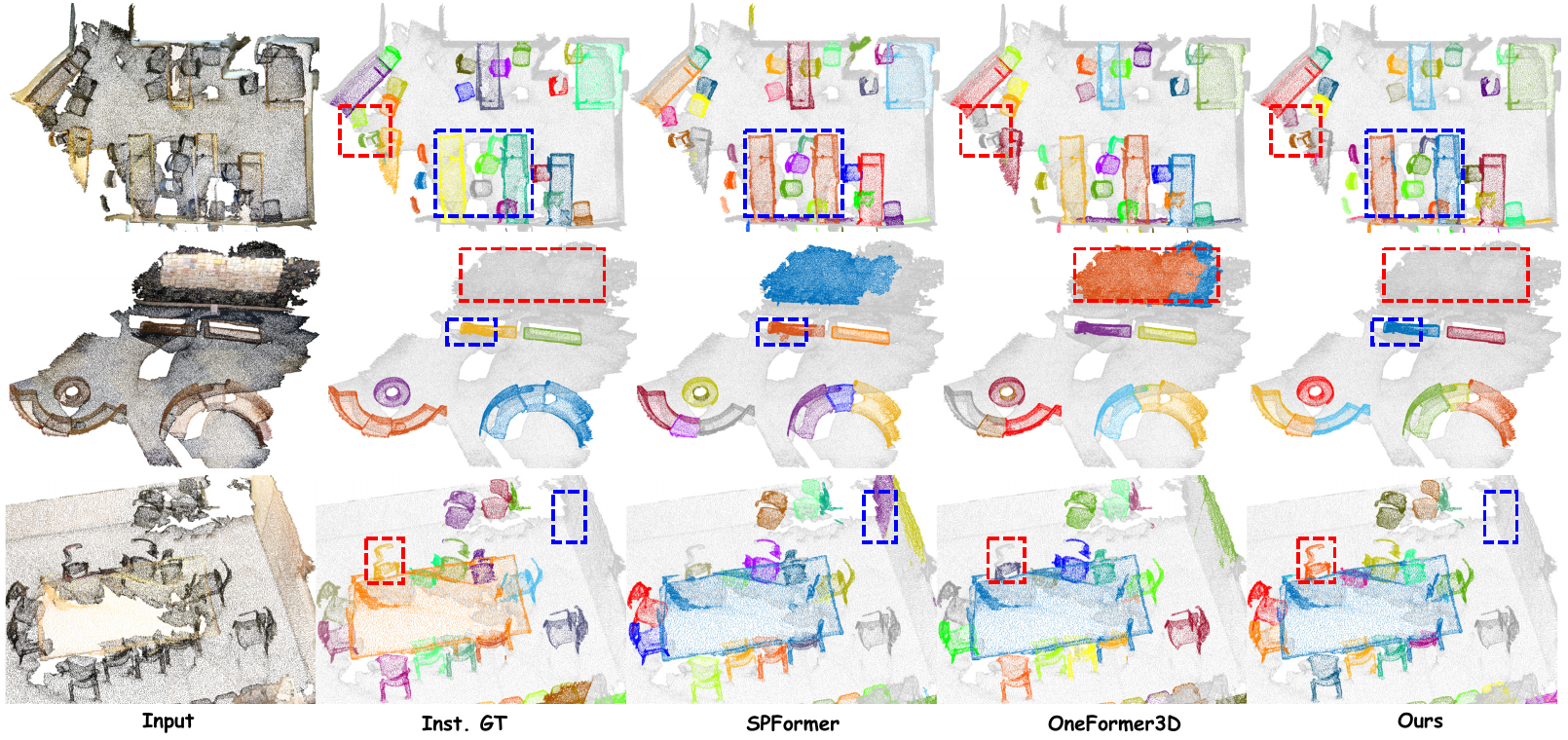}
    \caption{\textbf{Visualization results on \scannet validation set.} We compare visualization results of \ours (Ours) with SPFormer~\cite{sun2023spformer} and OneFormer3D~\cite{kolodiazhnyi2023of3d}. Input means the original point cloud, Inst. GT denotes ground truth instance masks, and different colors represent distinct instance IDs.}
    \label{fig:sup_viz_scannetv2}
\end{figure*}

\begin{figure}[t]
    \centering
    \includegraphics[width=0.47\textwidth]{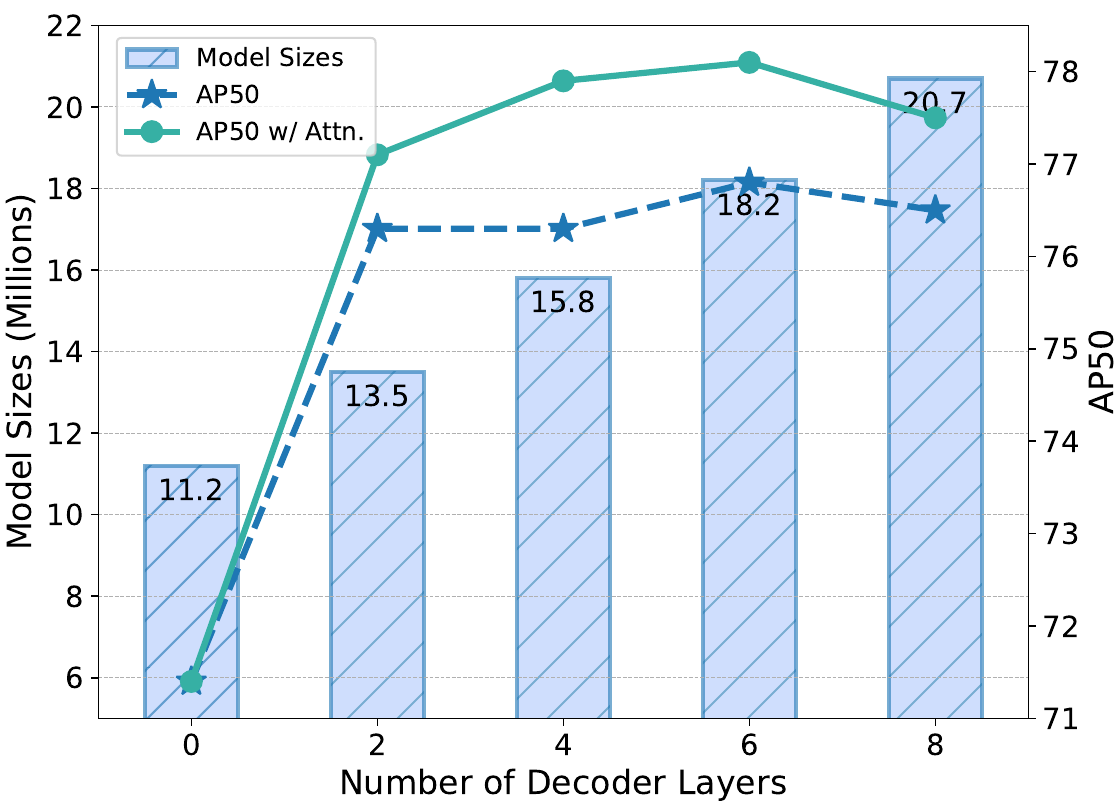}
    \caption{\textbf{Impact of stacked decoder layers $L$ for AP$_{50}$ and model size.}}
    \label{fig:decoder_layer_num}
\end{figure}

\subsection{Analysis and Ablation Study}
We conduct ablation studies on \scannet validation set to evaluate the effectiveness of our designs.

\begin{table}[t]
    \centering
    \caption{\textbf{Plug-and-play ability of our decoder.} $*$ indicates using Res16UNet34C as backbone.}
    \resizebox{0.49\textwidth}{!}{
    \begin{tabular}{y{23mm}y{8mm}y{8mm}y{8mm}y{12mm}}
        \toprule
        Methods & mAP & AP$_{50}$ & AP$_{25}$ & Memory$\downarrow$ \\
        \midrule
        \hspace{0.65em}SPFormer & 56.3 & 73.9 & 82.9 & 2014M \\
        \sm SPFormer & 55.4 & 74.6 & 83.8 & \textbf{1946M}$_{\textcolor{red}{\downarrow \textbf{68}}}$ \\
        \rowcolor{m_blue} \smpro SPFormer & \textbf{57.7}$_{\textcolor{red}{\textbf{+1.4}}}$ & \textbf{76.4}$_{\textcolor{red}{\textbf{+2.5}}}$ & \textbf{84.7}$_{\textcolor{red}{\textbf{+1.8}}}$ & 1992M$_{\textcolor{red}{\downarrow \textbf{22}}}$ \\
        \midrule
        \hspace{0.65em}Mask3D$^*$ & 55.2 & 73.7 & 83.5 & - \\
        \sm Mask3D & 56.2 & 76.1 & \textbf{84.2} & \textbf{1976M} \\
        \rowcolor{m_blue} \smpro Mask3D & \textbf{57.0}$_{\textcolor{red}{\textbf{+1.8}}}$ & \textbf{76.8}$_{\textcolor{red}{\textbf{+3.1}}}$ & \textbf{84.2}$_{\textcolor{red}{\textbf{+0.7}}}$ & 2030M \\
        \midrule
        \hspace{0.65em}OneFormer3D & 57.5 & 76.7 & 85.0 & 3038M \\
        \sm OneFormer3D & 57.1 & 75.5 & 83.8 & \textbf{2250M}$_{\textcolor{red}{\downarrow \textbf{788}}}$ \\
        \rowcolor{m_blue} \smpro OneFormer3D & \textbf{59.2}$_{\textcolor{red}{\textbf{+1.7}}}$ & \textbf{77.4}$_{\textcolor{red}{\textbf{+1.1}}}$ & \textbf{85.2}$_{\textcolor{red}{\textbf{+1.2}}}$ & 2560M$_{\textcolor{red}{\downarrow \textbf{478}}}$ \\
        \bottomrule
    \end{tabular}
}
    \label{tab:plug}
\end{table}

\mypara{Hierarchical semantic-spatial query initializer.} We first investigate the impact of our initializer in~\tabref{tab:query_decoder}. Specifically, we compare it with two query initialization strategies, FPS~\cite{schult2023mask3d} and semantic confidence-based selection~\cite{he2023fastinst}, maintaining the same query number. Quantitative results show that our initialization method achieves the best performance among them. We further visualize the initialized query distributions in~\figref{fig:teaser}a, which shows that our initializer effectively selects prominent queries with high semantic confidence and spatial distribution. In \figref{fig:query_convergence}, we compare the convergence speed. It is evident that our proposed method converges faster, indicating that it leads to high-quality queries.

\mypara{Coordinate-guided SSM query decoder.} In~\tabref{tab:query_decoder} \textit{bottom}, we explore the impact of our decoder by comparing it with four different efficient decoders: AdaMixer~\cite{gao2022adamixer}, KVT~\cite{wang2022kvt}, PEM~\cite{cavagnero2024pem}, and Linear Attention~\cite{katharopoulos2020transformers}, where the last three introduce advanced efficient attention mechanisms. We also report corresponding peak GPU memory consumption during inference to provide a comprehensive evaluation. Although KVT achieves comparable AP$_{50}$ and memory usage, it suffers from a notable performance drop in mAP. PEM and Linear Attention obtain similar accuracy with almost the same GPU consumption. However, our \ours outperforms them by 1.0\% and 0.9\% in AP$_{50}$, respectively, using less memory. AdaMixer shows inferior results in all metrics compared to our method. These comparisons reveal the superiority of our decoder in balancing performance and efficiency. In~\tabref{tab:components}, we analyze the effectiveness of core components in our decoder by replacing them with conventional attention mechanisms. Our dual-path SSM block attains better accuracy than self-attention when combined with masked cross-attention (\#2 \textit{v.s.} \#1), which is attributed to the query serialization for positional clues incorporation and causal relationship modeling of SSM. The local aggregation scheme also presents advantages over the cross-attention mechanism (\smpro Ours \textit{v.s.} \#2), regardless of accuracy or memory, as it can effectively capture local features and reduce redundant computation.

\begin{figure*}[!htbp]
    \centering
    \includegraphics[width=0.98\textwidth]{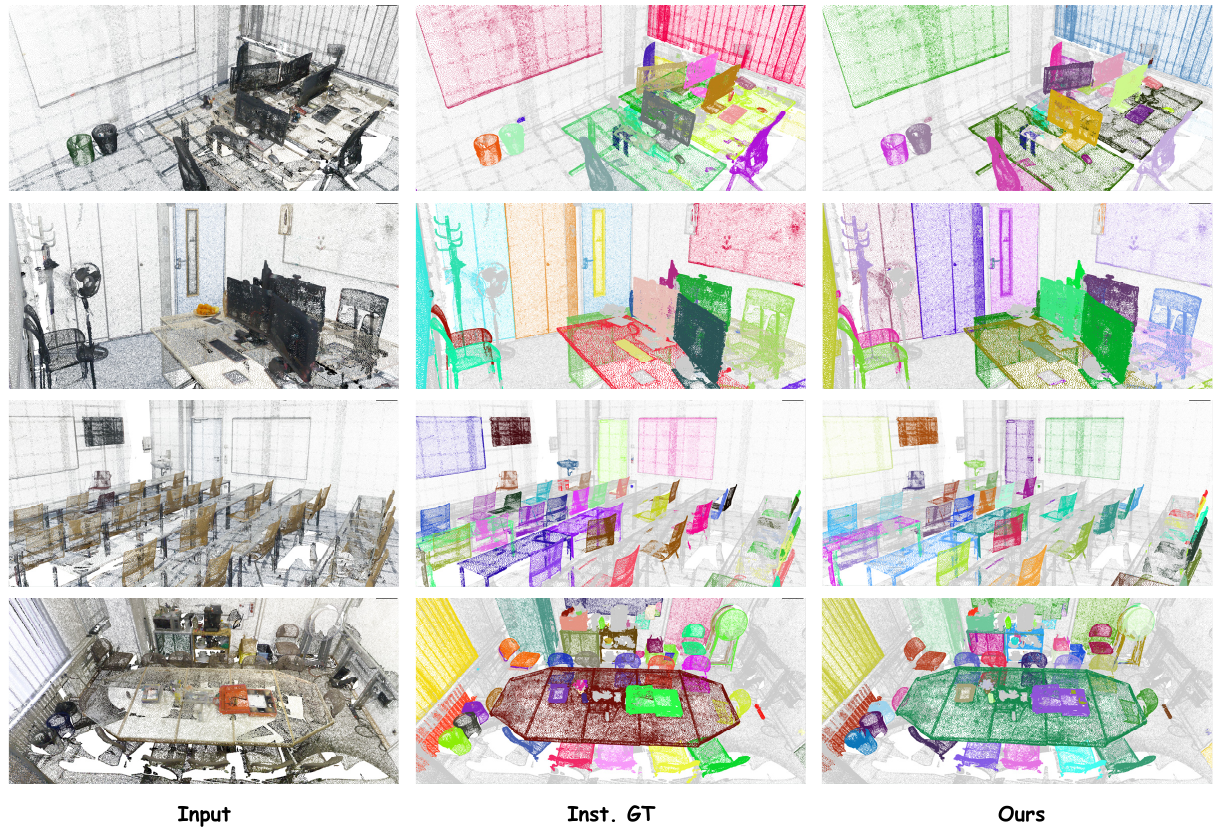}
    \caption{\textbf{Visualization results on \scannetpp V2 dataset.} Input, Inst. GT, and Ours indicate the input point cloud, instance ground truth, and predicted instance masks, respectively. Different colors are used to represent different instance IDs.}
    \label{fig:sup_viz_scannetpp}
\end{figure*}

\begin{table}[t]
    \centering
    \caption{\textbf{Local aggregation number $k$ and selection $r$.}}
    \resizebox{0.49\textwidth}{!}{
    \begin{tabular}{y{5mm}x{4mm}x{4mm}x{8mm}x{4mm}|x{4mm}x{8mm}x{4mm}x{4mm}}
        \toprule
        \multirow{2}{*}{Values} & \multicolumn{4}{c}{\textit{ratio} $r$} & \multicolumn{4}{c}{$k$} \\
        \cmidrule(lr){2-5} \cmidrule(lr){6-9}
        {} & 0.4 & 0.6 & \cellcolor{gray} 0.8 & 0.9 & 4 & \cellcolor{gray} 8 & 16 & 32 \\
        \midrule
        mAP & 57.3 & 57.8 & \cellcolor{gray} \textbf{58.4}$_{\textcolor{red}{\textbf{+0.3}}}$ & 58.1 & 57.6 & \cellcolor{gray} \textbf{58.4}$_{\textcolor{red}{\textbf{+0.8}}}$ & 56.9 & 57.6 \\
        AP$_{50}$ & 76.2 & 77.4 & \cellcolor{gray} \textbf{78.1}$_{\textcolor{red}{\textbf{+0.3}}}$ & 77.8 & 77.3 & \cellcolor{gray} \textbf{78.1}$_{\textcolor{red}{\textbf{+0.4}}}$ & 77.7 & 77.6 \\
        AP$_{25}$ & 83.8 & 85.6 & \cellcolor{gray} \textbf{86.1}$_{\textcolor{red}{\textbf{+0.4}}}$ & 85.7 & 84.8 & \cellcolor{gray} \textbf{86.1}$_{\textcolor{red}{\textbf{+0.0}}}$ & \textbf{86.1} & 85.9 \\
        \bottomrule
    \end{tabular}
}

    \label{tab:ablation_k}
\end{table}

\begin{table}[t]
    \centering
    \caption{\textbf{Impact of query numbers $q$.}}
    \resizebox{0.49\textwidth}{!}{
    \begin{tabular}{x{16mm}|x{7mm}x{7mm}x{9mm}x{7mm}x{7mm}}
        \toprule
        \textit{\#num query} & 100 & 200 & \cellcolor{gray} 400 & 600 & 800 \\
        \midrule
        mAP & 56.1 & 57.7 & \cellcolor{gray} \textbf{58.4}$_{\textcolor{red}{\textbf{+0.1}}}$ & 58.3 & 58.0 \\
        AP$_{50}$ & 76.0 & 77.1 & \cellcolor{gray} \textbf{78.1}$_{\textcolor{red}{\textbf{+0.3}}}$ & 77.8 & 77.5 \\
        AP$_{25}$ & 84.7 & 85.0 & \cellcolor{gray} \textbf{86.1}$_{\textcolor{red}{\textbf{+0.1}}}$ & 86.0 & 85.5 \\
        \bottomrule
    \end{tabular}
}
    \label{tab:ablation_query_num}
\end{table}

\begin{figure}[t]
    \centering
    \includegraphics[width=0.47\textwidth]{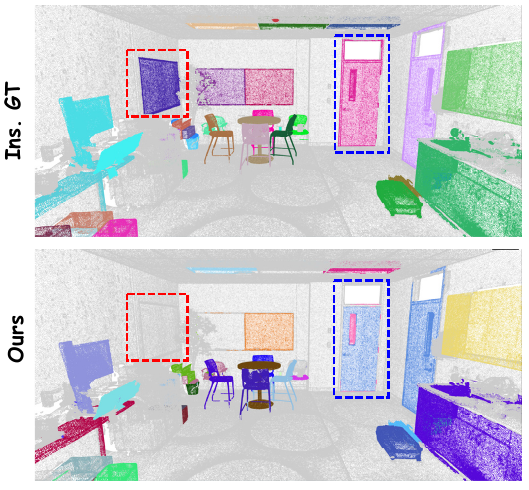}
    \caption{\textbf{Failure cases on \scannetpp V2 dataset.} The \textcolor{red}{red box} shows the failure case of the blackboard with a large convex shape and color similar to the background. The \textcolor{blue}{blue box} illustrates the excessive segmentation problem, where the door is incorrectly segmented into multiple instances.}
    \label{fig:sup_viz_failure}
\end{figure}

\begin{table}[t]
    \centering
    \caption{\textbf{Other ablation studies.} T. Hilbert and T. Z mean Trans. Hilbert order and Trans. Z-order, respectively. Local Aggreg. Layers refer to the local aggregation layers in the spatial dual-branch SSM decoder.}
    \resizebox{0.49\textwidth}{!}{
    \begin{tabular}{x{3mm}y{40mm}y{7mm}y{7mm}y{7mm}}
        \toprule
        \# & Variants & mAP & AP$_{50}$ & AP$_{25}$ \\
        \midrule
        \multicolumn{5}{c}{\textbf{\textit{Serialization Patterns}}} \\
        \cmidrule(lr){1-5}
        1 & Hilbert & 58.2 & 77.4 & 85.5 \\
        2 & T. Hilbert & 58.3 & 77.5 & 85.3 \\
        \rowcolor{gray} 3 & Hilbert + T. Hilbert & \textbf{58.4} & 78.1 & \textbf{86.1} \\
        4 & Z & 58.1 & 77.5 & 85.2 \\
        5 & Z + T. Z & 58.3 & 77.8 & 85.8 \\
        6 & Hilbert + T. Hilbert + Z + T. Z & 58.1 & \textbf{78.3} & 85.6 \\
        \midrule
        \multicolumn{5}{c}{\textbf{\textit{Query Coordinate Update}}} \\
        \cmidrule(lr){1-5}
        7 & \textit{w/o} Coordinate Update  & 56.5 & 77.0 & 84.7 \\
        \rowcolor{gray} 8 & \textit{w/ } Coordinate Update & \textbf{58.4} & \textbf{78.1} & \textbf{86.1} \\
        \midrule
        \multicolumn{5}{c}{\textbf{\textit{Local Aggregation}}} \\
        \cmidrule(lr){1-5}
        9 & \textit{w/o} Local Aggreg. Layers & 57.8 & 77.3 & 85.9 \\
        \rowcolor{gray} 10 & \textit{w/} Local Aggreg. Layers & \textbf{58.4} & \textbf{78.1} & \textbf{86.1} \\
        \bottomrule
    \end{tabular}
}
    \label{tab:others}
\end{table}

\mypara{Plug-and-play ability.} We assess the plug-and-play ability of our decoder in~\tabref{tab:plug}. We adapt it to several query-based methods by replacing their decoders, including SPFormer~\cite{sun2023spformer}, Mask3D~\cite{schult2023mask3d}, and OneFormer3D~\cite{kolodiazhnyi2023of3d}. The table shows that our decoder can be easily integrated into existing algorithms and obtain better performance with less memory consumption in most cases. These experiments indicate that our query decoder can facilitate efficiency without sacrificing performance.

\mypara{Selection ratio $r$ and local aggregation number $k$.} In~\tabref{tab:ablation_k} \textit{left}, we explore the influence of selection ratio $r$ in the initializer, ranging from 0.4 to 0.9. The results show that a selection ratio of 0.8 achieves the best performance.
The reception field of our local aggregation is determined by the number $k$, which in turn affects model accuracy. We evaluate its effect in~\tabref{tab:ablation_k} \textit{right}. When increasing $k$ from 4 to 8, metrics improve accordingly. However, further increasing $k$ leads to a slight performance drop, which is attributed to the introduction of noisy information.

\mypara{Stacked layers $L$ and query numbers $q$.} As shown in~\figref{fig:decoder_layer_num}, we analyze the AP$_{50}$ impact and model size of decoder layers by varying $L$ from 0 to 8 with a step size of 2. The results tends to stabilize at 6 layers for both \sm \ours (\raisebox{3\height}{\tikz \draw[dashed, thick, color=smap50] (0,0) -- (3mm,0);}) and \smpro \ours (\raisebox{3\height}{\tikz \draw[thick, color=smpro] (0,0) -- (3mm,0);}). Even without decoder layers, our method still achieves 71.4\% AP$_{50}$, showing its advantages in initializing queries. In~\tabref{tab:ablation_query_num}, we explore the effect of query numbers $q$ and find that performance improves with increasing $q$ from 100 to 400, but further increasing $q$ yields a slight decline, indicating that a moderate number is beneficial for the performance.

\mypara{Query serialization.} In \ours, we propose serializing the query set into ordered sequences to incorporate positional information and model causal relationships. In~\tabref{tab:others} \#1-\#}, we conduct experiments to evaluate different query serialization patterns. We test various variants, including Hilbert, Trans. Hilbert, Z-order, Trans. Z-order and their combinations. The comparison between \#1, \#2, and \#3 indicates the effectiveness of our proposed spatial dual-path SSM block in mitigating the limitations of unidirectional modeling. Our results show that the combination of Hilbert and Trans. Hilbert (\#3) outperforms Z-order counterparts, which is attributed to the better spatial locality of Hilbert order. Although combining all patterns (\#6) yields comparable performance, it also introduces extra computational costs.

\mypara{Query coordinate update.} As discussed in~\secref{subsec:pgqd}, we propose updating corresponding query coordinates during the refinement process. We argue that this design ensures more accurate query orders and boosts performance. To verify this, we conduct an ablation study in~\tabref{tab:others} \#7-\#8 by comparing the results with and without updates. The results clearly demonstrate that our design achieves better performance, with improvements of 1.9\% mAP and 1.1\% AP$_{50}$.

\mypara{Local aggregation layers.} In our hybrid decoder~\smpro, we leverage local aggregation layers in the latter layers. To validate its effectiveness, we conduct an ablation study in~\tabref{tab:others} \#9-\#10 by comparing the performance with and without them. It is evident that incorporating local aggregation layers leads to performance improvements, achieving 0.6\% mAP and 0.8\% AP$_{50}$ gains. This demonstrates its advantage in capturing local features and enhancing query refinement.

\subsection{Qualitative Results}
\mypara{Qualitative results on \scannet dataset.} In~\figref{fig:sup_viz_scannetv2}, we illustrate visualization results on \scannet validation set comparing with SPFormer~\cite{sun2023spformer} and OneFormer3D~\cite{kolodiazhnyi2023of3d}. For nearby instances that have similar shapes, our model can correctly separate them with fine-grained details, such as chairs (row 1 \textcolor{red}{red box}) and tables (row 1 \textcolor{blue}{blue box}). In contrast, both SPFormer and OneFormer3D fail to segment them as distinguished instances. Benefiting from the proposed hierarchical semantic-spatial query initializer, our framework can prevent the influence of noisy information. For example, in row 2 \textcolor{red}{red box} and row 3 \textcolor{blue}{blue box}, our method can avoid the interference of background, while SPFormer and OneFormer3D falsely predict background regions as instances.
Moreover, \ours can accurately recognize instance boundaries with complex shapes, such as sofas (row 2 \textcolor{blue}{blue box}) and chairs (row 3 \textcolor{red}{red box}).

\mypara{Qualitative results on the \scannetpp V2 dataset.} In~\figref{fig:sup_viz_scannetpp}, we present four examples on \scannetpp V2 validation split to demonstrate the superiority of our method in handling complex and high-fidelity scenes. Specifically, we show the input point cloud, instance ground truth, and predicted instance masks. 
We can observe from~\figref{fig:sup_viz_scannetpp} that \ours can accurately segment instances with complex shapes and layouts.

\mypara{Failure cases.} Despite our \ours achieving competitive performance on various datasets, it remains facing challenges in some special scenarios. In~\figref{fig:sup_viz_failure}, we show two failure cases on \scannetpp V2 validation split. In the \textcolor{red}{red box}, \ours fails to segment the blackboard with a large convex shape. This failure is likely due to the insufficient distinctive features between the blackboard and surrounding wall, making it difficult to accurately identify and separate the blackboard as a distinct instance. In the \textcolor{blue}{blue box}, our approach suffers from the excessive segmentation problem, where the door is incorrectly segmented into multiple instances. This over-segmentation error may be attributed to the model misinterpreting certain features or patterns on the door as boundaries between different objects, leading to the erroneous division of a single object into several parts. These failure cases indicate the limitations of our model and provide insights for future improvements.
\section{Conclusion}
In this paper, we introduce \ours, a novel framework leveraging a hierarchical semantic-spatial query initializer and a coordinate-guided SSM query decoder for 3D instance segmentation. Our method effectively alleviates the query initialization dilemma by considering both semantic cues and spatial distribution. Moreover, our proposed decoder achieves efficient query refinement and accurate instance prediction by utilizing a local aggregation scheme and a spatial dual-path SSM block, which captures underlying dependencies within the query set by considering significant positional information. This design enables the model to focus on geometrically coherent regions, avoiding the incorporation of noisy information. As a result, \ours ranks first place on the latest \scannetpp V2 leaderboard, outperforming the previous best method by 2.5\% mAP with only 1/3 FLOPs and obtains competitive performance on \scannet, \scannetp, \sdis, and \scannetpp V1 benchmarks with less computational cost. 
\section{Limitations and Future Work}
\label{sec:limitations}

Although \ours attains the balance between performance and efficiency, there are still some limitations and potential directions for future work. First, our method leverages same query number for all scenes, which may not be optimal since scenes have various scales. We will explore a more adaptive strategy to improve its flexibility. Furthermore, additional operations like Hilbert curve sorting is integrated to facilitate positional information capture, which may introduce additional processing time. We will investigate more straightforward techniques to reduce the latency. In addition, our current investigation is mainly focused on indoor scenes, and we plan to extend it to outdoor scenes, which have different characteristics compared to indoor scenes.

% \input{sections/6_appendix}

\section*{Acknowledgment}
The research work described in this paper was conducted in the JC STEM Lab of Machine Learning and Computer Vision funded by The Hong Kong Jockey Club Charities Trust. This research received partially support from the Global STEM Professorship Scheme from the Hong Kong Special Administrative Region.

\bibliographystyle{IEEEtran}
\bibliography{IEEEabrv, references}

\end{document}